\documentclass[letterpaper]{article} 
\usepackage{aaai25}  
\usepackage{times}  
\usepackage{helvet}  
\usepackage{courier}  
\usepackage[hyphens]{url}  
\usepackage{graphicx} 
\usepackage{natbib}  
\usepackage{caption} 
\usepackage{algorithm}
\usepackage{algorithmic}
\usepackage{booktabs}
\usepackage{multirow}
\usepackage{arydshln}
\usepackage[dvipsnames]{xcolor}
\usepackage{amssymb}
\usepackage{amsmath}

\newcommand{\blue}[1]{\textcolor{blue}{\textbf{#1}}}
\usepackage{newfloat}
\usepackage{listings}
\DeclareCaptionStyle{ruled}{labelfont=normalfont,labelsep=colon,strut=off} 
\floatstyle{ruled}
\newfloat{listing}{tb}{lst}{}
\floatname{listing}{Listing}
\title{\textit{SegFace}: Face Segmentation of Long-Tail Classes
}
\author{
    Kartik Narayan,
    Vibashan VS,
    Vishal M. Patel
}
\affiliations{
    Johns Hopkins University\\
    \{knaraya4, vvishnu2, vpatel36\}@jhu.edu \\
    {\textcolor{magenta}{\url{https://github.com/Kartik-3004/SegFace}}}
}

\begin{document}

\maketitle

\begin{abstract}
Face parsing refers to the semantic segmentation of human faces into key facial regions such as eyes, nose, hair, etc. It serves as a prerequisite for various advanced applications, including face editing, face swapping, and facial makeup, which often require segmentation masks for classes like eyeglasses, hats, earrings, and necklaces. These infrequently occurring classes are called long-tail classes, which are overshadowed by more frequently occurring classes known as head classes. Existing methods, primarily CNN-based, tend to be dominated by head classes during training, resulting in suboptimal representation for long-tail classes. Previous works have largely overlooked the problem of poor segmentation performance of long-tail classes. To address this issue, we propose \textit{SegFace}, a simple and efficient approach that uses a lightweight transformer-based model which utilizes learnable class-specific tokens. The transformer decoder leverages class-specific tokens, allowing each token to focus on its corresponding class, thereby enabling independent modeling of each class. The proposed approach improves the performance of long-tail classes, thereby boosting overall performance. To the best of our knowledge, \textit{SegFace} is the first work to employ transformer models for face parsing. Moreover, our approach can be adapted for low-compute edge devices, achieving $95.96$ FPS. We conduct extensive experiments demonstrating that \textit{SegFace} significantly outperforms previous state-of-the-art models, achieving a mean F1 score of $88.96$ ($+2.82$) on the CelebAMask-HQ dataset and $93.03$ ($+0.65$) on the LaPa dataset.
\end{abstract}    

\begin{figure}[t]
    \centering
    \includegraphics[width=0.95\linewidth]{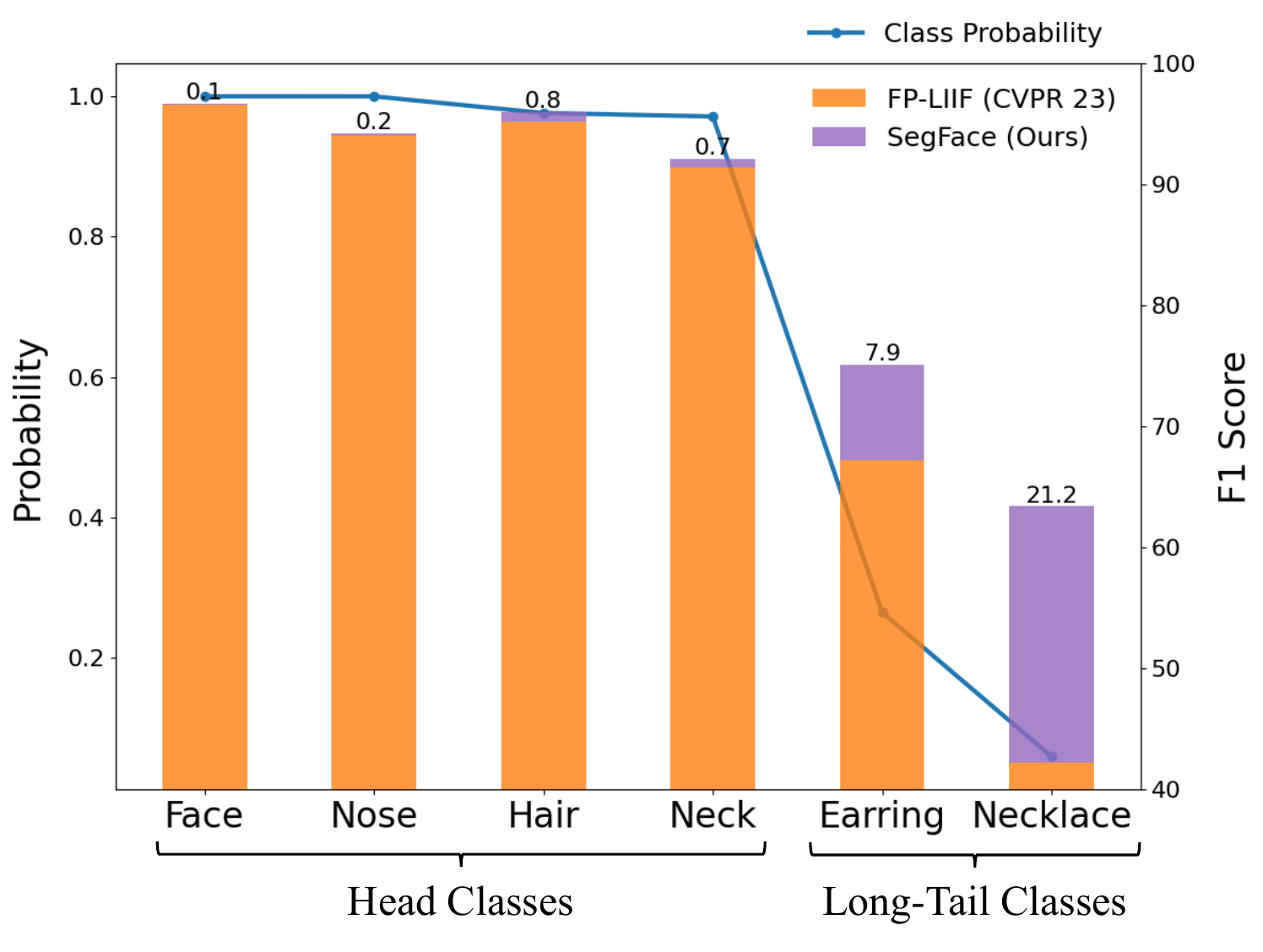}
    \caption{The proposed \textit{SegFace} leverages a lightweight transformer decoder with learnable class-specific tokens. The association of each class with a token enables the independent modeling of each class, which boosts the segmentation performance of long-tail classes that typically underperform in existing methods. The blue line represents the probability of a class being present in a randomly selected image from the CelebAMask-HQ train set. \textit{SegFace} provides a significant boost in the segmentation performance of long-tail classes ($+7.9$, $+21.2$), thereby establishing a new state-of-the-art in face parsing performance.}
    \label{fig:viz_abstract}
\end{figure}

\section{Introduction}
Face parsing, a semantic segmentation task, involves assigning pixel-level labels to a face image to distinguish key facial regions, such as the eyes, nose, hair, and ears. The identification of different facial regions is crucial for a variety of applications, including face swapping~\cite{xu2022region}, face editing~\cite{lee2020maskgan}, face generation~\cite{zhang2023adding}, face completion~\cite{li2017generative}, and facial makeup~\cite{wan2022facial}. Long-tail classes are those that occur infrequently within a dataset. Existing face parsing datasets~\cite{lee2020maskgan} consist of these long-tail classes, which are mostly accessories like eyeglasses, necklaces, hats, and earrings, because not all faces will feature these items. We cannot expect to have equal representation of all classes in current or even future face-parsing datasets, as certain facial attributes like hair, nose and eyes are naturally more common than accessories like earrings and necklaces. Additionally, it is difficult to collect samples with less frequently occurring classes. Moreover, detailed annotation for face segmentation, especially for less common or smaller facial features, is labor-intensive and costly.

Since the advent of deep learning in semantic segmentation~\cite{long2015fully}, numerous studies have focused on solving face segmentation. Several works~\cite{guo2018residual, zhou2015interlinked, lin2021roi} leverage the learning potential of deep convolutional neural networks to achieve promising face segmentation performance. AGRNet~\cite{te2021agrnet} introduces an adaptive graph representation approach that learns and reasons over facial components by representing each component as a vertex and relating each vertex, while also incorporating image edges as a prior to refine parsing results. Similarly, EAGRNet~\cite{te2020edge} extends this approach by enabling reasoning over non-local regions to capture global dependencies between distinct facial components. Recently, FaRL~\cite{zheng2022general} explored pre-training on a large image-text face dataset to enhance performance on downstream tasks, demonstrating that their pre-trained weights outperform those based on ImageNet~\cite{5206848}. DML-CSR~\cite{zheng2022decoupled} utilizes a multi-task model for face parsing, edge detection, and category edge detection, incorporating a dynamic dual graph convolutional network to address spatial inconsistency and cyclic self-regulation for noisy labels. The recent FP-LIIF~\cite{Sarkar_2023_CVPR} leverages the structural consistency of the human face using a lightweight Local Implicit Function Network with a simple convolutional encoder-pixel decoder architecture, notable for its small parameter size and high FPS, making it ideal for low-compute devices.  Despite these advancements, most prior works have focused on specific challenges, such as improving the correlation between facial components, enhancing hair segmentation, handling noisy labels, and optimizing inference speed. However, they often neglect the critical issue of long-tail class performance, leading to suboptimal results in long-tail classes (see Figure~\ref{fig:viz_abstract}).

To overcome this issue, we propose \textit{SegFace}, a systematic approach that enhances the segmentation performance of long-tail classes. These classes are often underrepresented in the dataset, typically including accessories like earring and necklace, while head classes are more frequent and include regions like the face and hair. In a face image, regions like the eyes, mouth, and accessories (long-tail classes) are naturally smaller than the overall face and hair regions (head classes). Using only the final single-scale feature of a model for face segmentation can lead to a loss of detail, as facial features appear at different scales. Our approach leverages a Swin Transformer backbone to extract features at multiple scales, helping to mitigate the scale discrepancy between different face regions. Multi-scale feature extraction effectively captures both fine details and larger structures, aiding the model in capturing the global context of the face. We fuse the multi-scale features using MLP fusion to obtain the fused features, which are then input to the \textit{SegFace} decoder. The lightweight transformer decoder utilizes learnable class-specific tokens, each associated with a particular class. We employ cross-attention between the fused features and learnable tokens, enabling each token to extract class-specific information from the fused features. This design allows the tokens to focus specifically on their corresponding classes, promoting independent modeling of all classes and mitigating the problem of dominant head classes overshadowing long-tail classes during training. 

The key contributions of our work are as follows: 
\begin{itemize}
    \item We introduce a lightweight transformer decoder with learnable class-specific tokens, that ensures each token is dedicated to a specific class, thereby enabling independent modeling of classes. The design effectively addresses the challenge of poor segmentation performance of long-tail classes, prevalent in existing methods.
    \item Our multi-scale feature extraction and MLP fusion strategy, combined with a transformer decoder that leverages learnable class-specific tokens, mitigates the dominance of head classes during training and enhances the feature representation of long-tail classes.
    \item \textit{SegFace} establishes a new state-of-the-art performance on the LaPa dataset (93.03 mean F1 score) and the CelebAMask-HQ dataset (88.96 mean F1 score). Moreover, our model can be adapted for fast inference by simply swapping the backbone with a MobileNetV3 backbone. The mobile version achieves a mean F1 score of 87.91 on the CelebAMask-HQ dataset with 95.96 FPS.  
\end{itemize}
\section{Related Work}
\subsection{Face Parsing}
Early face parsing approaches employed techniques such as exemplars~\cite{smith2013exemplar}, probabilistic index maps~\cite{scheffler2011joint}, Gabor filters~\cite{hernandez2015facial}, and low-rank decomposition~\cite{guo2015facial}. Since the rise of deep learning, numerous deep convolutional network-based methods have been proposed for face segmentation~\cite{warrell2009labelfaces, khan2015multi, liang2015deep, lin2019face, liu2017face}. Recently, AGRNet~\cite{te2021agrnet} and EAGRNet~\cite{te2020edge} proposed graph representation-based methods that correlate different facial components and utilize edge information for parsing. DML-CSR~\cite{zheng2022decoupled} explores multi-task learning and introduces a dynamic dual graph convolutional network to address spatial inconsistency and cyclic self-regulation to tackle the presence of noisy labels. Local-based methods, which are most similar to our work, aim to predict each facial part individually by training separate models for different facial regions.~\cite{luo2012hierarchical} leverages a hierarichal approach to parse each component separately, while~\cite{zhou2015interlinked} propose using multiple CNNs that take input at different scales, fusing them through an interlinking layer that efficiently integrates local and contextual information. However, existing local-based approaches fail to benefit from a shared backbone and joint optimization, leading to suboptimal performance. \textit{SegFace} addresses this issue by independently modeling all the classes using learnable class-specific tokens, while still benefiting from multi-scale fused features extracted from a shared backbone.

\begin{figure*}[t]
    \centering
    \includegraphics[width=0.98\textwidth]{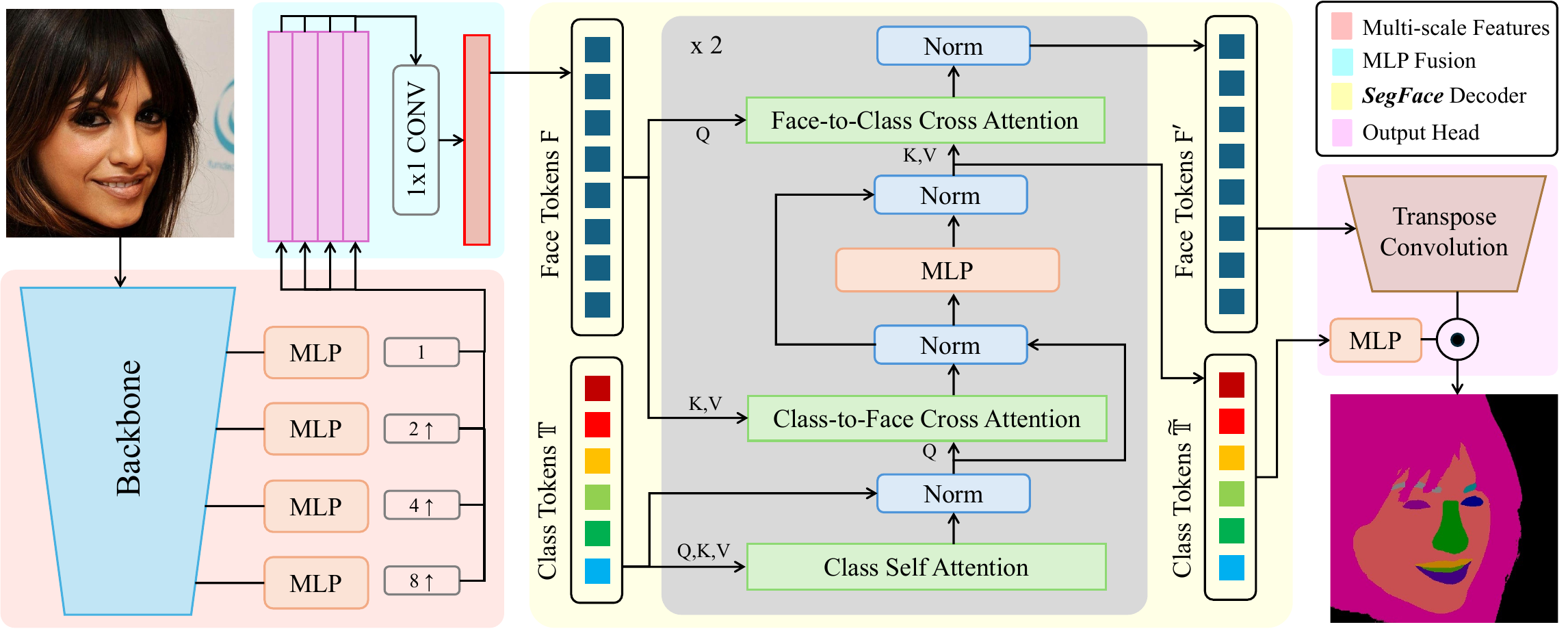}
    \caption{The proposed architecture, \textit{SegFace}, addresses face segmentation by enhancing the performance on long-tail classes through a transformer-based approach. Specifically, multi-scale features are first extracted from an image encoder and then fused using an MLP fusion module to form face tokens. These tokens, along with class-specific tokens, undergo self-attention, face-to-token, and token-to-face cross-attention operations, refining both class and face tokens to enhance class-specific features. Finally, the upscaled face tokens and learned class tokens are combined to produce segmentation maps for each facial region.}
    \label{fig:archi}
\end{figure*}
\subsection{Transformers}
Transformer-based models such as ViT \cite{dosovitskiy2020image} and DETR \cite{carion2020end} have demonstrated their effectiveness in segmentation tasks by leveraging attention mechanisms to capture long-range dependencies and global context within images. Segformer \cite{xie2021segformer} and SETR \cite{zheng2021rethinking} are notable works which have shown that transformers can outperform traditional CNNs in general segmentation tasks. However, the application of transformers in face segmentation remains relatively underexplored, despite their potential advantages. Face segmentation presents unique challenges, such as the need for precise boundary detection and sensitivity to subtle variations in facial features, which traditional CNNs have addressed effectively. However, recent transformer-based segmentation networks like Mask2Former \cite{cheng2022masked} and SAM \cite{kirillov2023segment} have shown promising results in capturing both global and fine-grained contexts, leading to more accurate segmentation. These models leverage self-attention and cross-attention mechanisms, which can be viewed as non-local mean operations that compute the weighted average of all inputs. As a result, each class's inputs are calculated independently and averaged, allowing the model to selectively attend to relevant features without spatial constraints. This leads to a richer, contextualized representation, which can significantly benefit the understanding of long-tail visual relationships.
\section{Proposed Work}
The human face consists of various regions, including the nose, eyes, mouth, and accessories like earrings and necklaces. In face segmentation, these regions are treated as different classes, which vary in scale and frequency of occurrence. Classes such as hair and nose, naturally appear more often in a face image and are referred to as head classes. In contrast, accessories, which may not be present in every face image, are called long-tail classes and are underrepresented in face segmentation datasets. We calculate the frequency of each class in the dataset and determine the probability of a class occurring in a face image of the CelebAMask-HQ dataset. As shown in Figure~\ref{fig:viz_abstract}, the probability of a head class being present in an image is approximately $1.0$, compared to $0.26$ and $0.05$ for long-tail classes. Upon analyzing current face segmentation methods, we observe that they often perform poorly on long-tail classes. Our goal is to enhance the segmentation performance of long-tail classes, thereby boosting overall face segmentation performance.

Given a batch of face images \(I \in \mathbb{R}^{B \times H \times W \times 3}\), consisting of $N$ classes, where $B$ is the batch size, while $H$ and $W$ denote the height and width of the image, respectively. \textit{SegFace} extracts multi-scale features \(\mathbb{G} = \{G_i | 1\leq i\leq4\}\) from the intermediate layers of the image encoder $E_{\theta}$. These features are then fused using a MLP fusion module $f_{\phi}$ to obtain the face tokens $F$. The face tokens, along with their corresponding positional encodings, and the learnable class-specific tokens $\mathbb{T} = \{T_i | 1 \leq i \leq N\}$, are processed by the light-weight \textit{SegFace} decoder $g_{\psi}$ through self-attention and cross-attention operations, resulting in the learned class tokens $\mathbb{\tilde{T}}$ and updated face tokens $F'$. The updated face tokens are then upscaled using an upscaling module $h_{\alpha}$ and multiplied element-wise with the learned class tokens $\mathbb{\tilde{T}}$ after the tokens has been passed through an MLP to obtain the final segmentation map \(\mathbb{S} = \{S_i|1 \leq i\leq N\}\) where \(S_i\in \mathbb{R}^{B \times 1 \times H \times W}\), represents the segmentation map for each class. The complete process is as follows:
\[
\mathbb{\tilde{T}}', F' = g_{\psi}(F, \mathbb{T}), \quad \text{ where } F=f_{\phi}(E_{\theta}(I))
\]
\[
S_i = h_{\alpha}(F') \odot \text{MLP}(\mathbb{\tilde{T}}_{i})
\]
Here, $S_i$ is the output segmentation map for the $i$-th class. We utilize these segmentation maps to calculate the loss. We use cross entropy loss along with dice loss to train the complete pipeline which is illustrated in Figure~\ref{fig:archi}. The final loss function can be given as: $L = \lambda_1L_{\text{dice}} + \lambda_2L_{\text{CE}}$.

\subsection{Multi-scale Feature Extraction}
We perform multi-scale feature extraction to address the problem of scale discrepancy between different face regions. This approach effectively captures both fine details and larger structures, helping to obtain a comprehensive global context of the face and better handle the varying sizes and shapes of facial components. The multi-scale features are extracted from the image encoder $E_{\theta}$.
Let the batch of input images be \(I \in \mathbb{R}^{B \times H \times W \times 3}\), where \(B\) is the batch size, and \(H\) and \(W\) are the height and width of the image. The encoder extracts features from multiple layers:
\[
\mathbb{G} = \{G_i \mid 1 \leq i \leq 4\}, \quad G_i \in \mathbb{R}^{B \times C_i \times H_i \times W_i}
\]
Here, \(G_i\) represents the feature map extracted from the \(i\)-th layer of the encoder, \(C_i\) is the number of channels in the \(i\)-th feature map, and \(H_i\) and \(W_i\) denote the height and width of the \(i\)-th feature map, respectively. The hierarchical features extracted from the encoder help capture coarse to fine-grained representations, making them suitable for segmenting smaller classes, which are often long-tail classes.

\subsection{MLP Fusion} 
We perform multi-scale feature aggregation using an MLP fusion module $f_{\theta}$ to obtain the face tokens that will be passed to the \textit{SegFace} decoder. In this module, the multi-scale features \(\mathbb{G} = \{G_i | 1\leq i\leq4\}\) are processed by separate MLPs, each corresponding to a different scale, to make the channel dimension consistent for fusion. Each MLP transforms its corresponding \(G_i\) into a feature map \(G_i'\) with a uniform number of channels \(C'\), as follows: \( G_i' = \text{MLP}_i(G_i), \text{ where } G_i' \in \mathbb{R}^{B \times C' \times H_i \times W_i} \). The resulting feature maps \(G_i'\) are then upsampled to match the spatial resolution of the first feature map \(G_1'\) using bilinear interpolation, represented as \( G_i'' = \text{Interp}(G_i'), \text{ where } G_i'' \in \mathbb{R}^{B \times C' \times H_1 \times W_1}, \forall i \in \{1, 2, 3, 4\} \). These upsampled multi-scale features \(G_i''\) are concatenated along the channel dimension to form a unified feature map. Finally, this concatenated feature map is passed through a single convolutional layer, to reduce the channel dimensionality back to \(C'\).
\[
F_{\text{concat}} = \text{Concat}(G_1'', G_2'', G_3'', G_4'') \in \mathbb{R}^{B \times (4 \times C') \times H_1 \times W_1}
\]
\[
F = \text{Conv1x1}(F_{\text{concat}}) \in \mathbb{R}^{B \times C' \times H_1 \times W_1}
\]
This fused feature map \(F\) represents the final multi-scale face tokens which is given as input to the \textit{SegFace} decoder.

\subsection{\textit{SegFace} Decoder}
The SegFace decoder is designed to model each class independently while enabling interactions between them, using learnable class-specific tokens. Let $\mathbb{T} = {T_i \in \mathbb{R}^{1 \times D} \mid 1 \leq i \leq N}$ represent these tokens, where $N$ is the number of classes, and $D$ is the embedding dimension (here, $D = 256$). These tokens are appended with positional encodings and correspond to various facial components, such as the background, face, eyes, nose, and other features. The decoder comprises of three main components: 1) Class-token Self-Attention, 2) Class-token to Face-token Cross-Attention, and 3) Face-token to Class-token Cross-Attention. Through self-attention and cross-attention operations within the transformer decoder, the tokens are guided to focus on class-specific features and facilitate interaction among different facial regions.

\textbf{Class-token Self-Attention:} This component facilitates interaction between different regions of the face by allowing each class token, $T_i$, to attend to all other class tokens. For each class token $T_i$, the operation is defined as:
\[
T'_i = \text{SelfAttention}(Q=T_i, K=\mathbb{T}, V=\mathbb{T}),
\]
where SelfAttention denotes the multi-head self-attention operation, and $Q$, $K$, and $V$ represent the queries, keys, and values, respectively. Each class token corresponds to a specific class, and the SelfAttention operation enables the model to learn the correlations between the structure and position of different facial regions.

\textbf{Class-token to Face-token Cross-Attention:} In this component, each class token $T'_i$ attends to the fused face token $F$, facilitating the extraction of class-specific information and enabling independent modeling of the classes. The updated class token $\tilde{T}_i$ is computed as follows:
\[
\tilde{T}_i = \text{CrossAttention}(Q=T'_i, K=F, V=F),
\]
where CrossAttention denotes the cross-attention operation. This mechanism ensures that long-tail classes are not overshadowed during training, as each class is associated with a token that extracts relevant features specifically for segmenting that long-tail class.

\textbf{Face-token to Class-token Cross-Attention:} In this component, the fused face tokens attend back to the learned class tokens, refining the face representation with class-specific information. The refined face token $F'$ is computed as follows:
\[
{F'} = \text{CrossAttention}(Q=F, K=\mathbb{\tilde{T}}, V=\mathbb{\tilde{T}})
\]
This component guides the feature extraction and fusion modules by aligning their training to ensure that the extracted features are enriched with class-specific information. 

\setlength\dashlinedash{2pt}
\setlength\dashlinegap{3pt}
\setlength\arrayrulewidth{0.4pt}

\begin{table*}[t]
\centering
\resizebox{\textwidth}{!}{
\renewcommand{\arraystretch}{1.15}
\begin{tabular}{|l|c|c|cccccccccc|c|c|}
    \hline
    \multicolumn{1}{|l|}{Method} & \multicolumn{1}{c|}{Venue}  & \multicolumn{1}{c|}{Resolution} & \multicolumn{1}{c}{Skin} & \multicolumn{1}{c}{Hair} & \multicolumn{1}{c}{Nose} & \multicolumn{1}{c}{L-Eye} & \multicolumn{1}{c}{R-Eye} & \multicolumn{1}{c}{L-Brow} & \multicolumn{1}{c}{R-Brow} & \multicolumn{1}{c}{L-Lip} & \multicolumn{1}{c}{I-Mouth} & \multicolumn{1}{c|}{U-Lip} & Mean F1 $\uparrow$ & Mean IoU $\uparrow$
    \\ \hline
    \multicolumn{1}{|l|}{Wei et al.} & \multicolumn{1}{c|}{TIP 19} & \multicolumn{1}{c|}{$512$} & $96.1$ & $95.1$ & $96.1$ & $88.9$ & $87.5$ & $86.0$ & $87.8$ & $83.8$ & $89.2$ & $83.1$ & $89.36$ & - 
    \\ 
    \multicolumn{1}{|l|}{BASS} & \multicolumn{1}{c|}{AAAI 20} & \multicolumn{1}{c|}{$473$} & $97.2$ & $96.3$ & $95.5$ & $88.1$ & $88.0$ & $87.7$ & $87.6$ & $85.7$ & $87.6$ & $84.4$ & $89.81$ & - 
    \\ 
    \multicolumn{1}{|l|}{EAGRNet} & \multicolumn{1}{c|}{ECCV 20} & \multicolumn{1}{c|}{$473$} & $97.3$ & $96.2$ & $97.1$ & $89.5$ & $90.0$ & $86.5$ & $87.0$ & $89.0$ & $90.0$ & $88.1$ & $91.07$ & - 
    \\ 
    \multicolumn{1}{|l|}{AGRNet} & \multicolumn{1}{c|}{TIP 21} & \multicolumn{1}{c|}{$473$} & $97.7$ & $96.5$ & $97.3$ & $91.6$ & $91.1$ & $89.9$ & $90.0$ & $90.1$ & $90.7$ & $88.5$ & $92.34$ & - 
    \\ 
    \multicolumn{1}{|l|}{FaRL\textsubscript{scratch}} & \multicolumn{1}{c|}{CVPR 22} & \multicolumn{1}{c|}{$512$} & $97.2$ & $93.1$ & $97.3$ & $91.6$ & $91.5$ & $90.1$ & $89.7$ & $89.1$ & $89.4$ & $87.2$ & $91.62$ & - 
    \\ 
    \multicolumn{1}{|l|}{DML-CSR} & \multicolumn{1}{c|}{CVPR 22} & \multicolumn{1}{c|}{$473$} & $97.6$ & $\blue{96.4}$ & $97.3$ & $91.8$ & $91.5$ & $90.4$ & $90.4$ & $89.9$ & $90.5$ & $88.0$ & $92.38$ & $87.13$
    \\ 
    \multicolumn{1}{|l|}{FP-LIIF} & \multicolumn{1}{c|}{CVPR 23} & \multicolumn{1}{c|}{$512$} & $97.5$ & $95.9$ & $97.2$ & $92.0$ & $92.2$ & $90.9$ & $90.6$ & $89.5$ & $90.3$ & $87.7$ & $92.38$ & - 
    \\ \hline
    \multicolumn{1}{|l|}{\textbf{\textit{SegFace}}} & \multicolumn{1}{c|}{AAAI 25} & \multicolumn{1}{c|}{$224$} & $97.5$ & $95.4$ & $97.3$ & $91.9$ & $92.1$ & $90.9$ & $90.8$ & $89.9$ & $90.8$ & $88.3$ & $92.50$ & $87.26$
    \\ 
    \multicolumn{1}{|l|}{\textbf{\textit{SegFace}}} & \multicolumn{1}{c|}{AAAI 25} & \multicolumn{1}{c|}{$256$} & $97.5$ & $95.7$ & $97.3$ & $92.2$ & $92.2$ & $91.0$ & $90.8$ & $90.0$ & $91.0$ & $88.4$ & $92.61$ & $87.45$
    \\ 
    \multicolumn{1}{|l|}{\textbf{\textit{SegFace}}} & \multicolumn{1}{c|}{AAAI 25} & \multicolumn{1}{c|}{$448$} & $97.7$ & $96.2$ & $97.5$ & $92.6$ & $92.7$ & $91.6$ & $91.4$ & $90.5$ & $91.4$ & $88.8$ & $93.03$ & $88.13$
    \\ 
    \multicolumn{1}{|l|}{\textbf{\textit{SegFace}}} & \multicolumn{1}{c|}{AAAI 25} & \multicolumn{1}{c|}{$512$} & $\blue{97.7}$ & $96.3$ & $\blue{97.5}$ & $\blue{92.6}$ & $\blue{92.7}$ & $\blue{91.6}$ & $\blue{91.4}$ & $\blue{90.5}$ & $\blue{91.2}$ & $\blue{88.7}$ & $\blue{93.03}$ & $\blue{88.14}$
    \\ \hline

\end{tabular}
}
\\ (a) LaPa Dataset
\\\textcolor{white}{a}\\
\resizebox{\textwidth}{!}{
\renewcommand{\arraystretch}{1.15}
\begin{tabular}{|l|c|c|ccccccccc|c|c|}
    \hline
    \multicolumn{1}{|l|}{\multirow{2}{*}{Method}} & \multicolumn{1}{c|}{\multirow{2}{*}{Venue}}  & \multicolumn{1}{c|}{\multirow{2}{*}{Resolution}} & \multicolumn{1}{c}{Face} & \multicolumn{1}{c}{Nose} & \multicolumn{1}{c}{E-Glasses} & \multicolumn{1}{c}{L-Eye} & \multicolumn{1}{c}{R-Eye} & \multicolumn{1}{c}{L-Brow} & \multicolumn{1}{c}{R-Brow} & \multicolumn{1}{c}{L-Ear} & \multicolumn{1}{c|}{R-Ear} & \multirow{2}{*}{Mean F1 $\uparrow$} & \multirow{2}{*}{Mean IoU $\uparrow$} \\
    \multicolumn{1}{|c|}{} & & & \multicolumn{1}{c}{I-Mouth} & \multicolumn{1}{c}{U-Lip} & \multicolumn{1}{c}{L-Lip} & \multicolumn{1}{c}{Hair} & \multicolumn{1}{c}{Hat} & \multicolumn{1}{c}{Earring} & \multicolumn{1}{c}{Necklace} & \multicolumn{1}{c}{Neck} & \multicolumn{1}{c|}{Cloth} & & 
    \\ \hline
    \multicolumn{1}{|l|}{\multirow{2}{*}{Wei et al.}} & \multicolumn{1}{|c|}{\multirow{2}{*}{TIP 19}} & \multicolumn{1}{|c|}{\multirow{2}{*}{$512$}} & $96.4$ & $91.9$ & $89.5$ & $87.1$ & $85.0$ & $80.8$ & $82.5$ & $84.1$ & $83.3$ & \multicolumn{1}{|c|}{\multirow{2}{*}{$82.06$}} & \multicolumn{1}{|c|}{\multirow{2}{*}{-}} \\ & & & $90.6$ & $87.9$ & $91.0$ & $91.1$ & $83.9$ & $65.4$ & $17.8$ & $88.1$ & $80.6$ & & 
    \\ \hdashline 
    \multicolumn{1}{|l|}{\multirow{2}{*}{EAGRNet}} & \multicolumn{1}{|c|}{\multirow{2}{*}{ECCV 20}} & \multicolumn{1}{|c|}{\multirow{2}{*}{$473$}} & $96.2$ & $94.0$ & $92.3$ & $88.6$ & $89.0$ & $85.7$ & $85.2$ & $88.0$ & $85.7$ & \multicolumn{1}{|c|}{\multirow{2}{*}{$84.89$}} & \multicolumn{1}{|c|}{\multirow{2}{*}{-}} \\ & & & $95.0$ & $88.9$ & $91.2$ & $94.9$ & $82.7$ & $68.3$ & $27.6$ & $89.4$ & $85.3$ & & 
    \\ \hdashline
    \multicolumn{1}{|l|}{\multirow{2}{*}{AGRNet}} & \multicolumn{1}{|c|}{\multirow{2}{*}{TIP 21}} & \multicolumn{1}{|c|}{\multirow{2}{*}{$473$}} & $96.5$ & $93.9$ & $91.8$ & $88.7$ & $89.1$ & $85.5$ & $85.6$ & $88.1$ & $88.7$ & \multicolumn{1}{|c|}{\multirow{2}{*}{$85.12$}} & \multicolumn{1}{|c|}{\multirow{2}{*}{-}} \\ & & & $92.0$ & $89.1$ & $91.1$ & $87.6$ & $87.2$ & $69.6$ & $32.8$ & $89.9$ & $84.9$ & & 
    \\ \hdashline
    \multicolumn{1}{|l|}{\multirow{2}{*}{FaRL\textsubscript{scratch}}} & \multicolumn{1}{|c|}{\multirow{2}{*}{CVPR 22}} & \multicolumn{1}{|c|}{\multirow{2}{*}{$512$}} & $96.2$ & $93.8$ & $92.3$ & $89.0$ & $89.0$ & $85.3$ & $85.4$ & $86.9$ & $87.3$ & \multicolumn{1}{|c|}{\multirow{2}{*}{$84.77$}} & \multicolumn{1}{|c|}{\multirow{2}{*}{-}} \\ & & & $91.7$ & $88.1$ & $90.0$ & $94.9$ & $82.7$ & $63.1$ & $33.5$ & $90.8$ & $85.9$ & & 
    \\ \hdashline
    \multicolumn{1}{|l|}{\multirow{2}{*}{DML-CSR}} & \multicolumn{1}{|c|}{\multirow{2}{*}{CVPR 22}} & \multicolumn{1}{|c|}{\multirow{2}{*}{$473$}} & $95.7$ & $93.9$ & $92.6$ & $89.4$ & $89.6$ & $85.5$ & $85.7$ & $88.3$ & $88.2$ & \multicolumn{1}{|c|}{\multirow{2}{*}{$86.07$}} & \multicolumn{1}{|c|}{\multirow{2}{*}{$77.81$}} \\ & & & $91.8$ & $89.1$ & $91.0$ & $94.5$ & $88.5$ & $69.6$ & $40.6$ & $89.6$ & $85.7$ & & 
    \\ \hdashline
    \multicolumn{1}{|l|}{\multirow{2}{*}{FP-LIIF}} & \multicolumn{1}{|c|}{\multirow{2}{*}{CVPR 23}} & \multicolumn{1}{|c|}{\multirow{2}{*}{$512$}} & $96.6$ & $94.0$ & $92.5$ & $90.0$ & $90.1$ & $85.6$ & $85.4$ & $86.8$ & $86.7$ & \multicolumn{1}{|c|}{\multirow{2}{*}{$86.14$}} & \multicolumn{1}{|c|}{\multirow{2}{*}{-}} \\ & & & $92.7$ & $89.4$ & $91.3$ & $95.2$ & $86.7$ & $67.2$ & $42.2$ & $91.4$ & $86.8$ & & 
    \\ \hline
    \multicolumn{1}{|l|}{\multirow{2}{*}{\textbf{\textit{SegFace}}}} & \multicolumn{1}{|c|}{\multirow{2}{*}{AAAI 25}} & \multicolumn{1}{|c|}{\multirow{2}{*}{$224$}} & $96.4$ & $93.8$ & $94.0$ & $90.1$ & $90.2$ & $86.0$ & $86.0$ & $88.2$ & $87.5$ & \multicolumn{1}{|c|}{\multirow{2}{*}{$87.47$}} & \multicolumn{1}{|c|}{\multirow{2}{*}{$79.65$}} \\ & & & $92.2$ & $89.4$ & $90.7$ & $95.7$ & $89.6$ & $71.1$ & $52.6$ & $91.5$ & $89.5$ & & 
    \\ \hdashline
    \multicolumn{1}{|l|}{\multirow{2}{*}{\textbf{\textit{SegFace}}}} & \multicolumn{1}{|c|}{\multirow{2}{*}{AAAI 25}} & \multicolumn{1}{|c|}{\multirow{2}{*}{$256$}} & $96.5$ & $93.9$ & $94.3$ & $90.2$ & $90.5$ & $86.3$ & $86.4$ & $88.5$ & $88.0$ & \multicolumn{1}{|c|}{\multirow{2}{*}{$87.66$}} & \multicolumn{1}{|c|}{\multirow{2}{*}{$79.91$}} \\ & & & $92.4$ & $89.6$ & $90.9$ & $95.8$ & $89.7$ & $72.0$ & $52.8$ & $91.5$ & $88.7$ & & 
    \\ \hdashline
    \multicolumn{1}{|l|}{\multirow{2}{*}{\textbf{\textit{SegFace}}}} & \multicolumn{1}{|c|}{\multirow{2}{*}{AAAI 25}} & \multicolumn{1}{|c|}{\multirow{2}{*}{$448$}} & $96.6$ & $94.1$ & $95.0$ & $90.8$ & $90.9$ & $87.0$ & $86.9$ & $89.2$ & $88.6$ & \multicolumn{1}{|c|}{\multirow{2}{*}{$88.77$}} & \multicolumn{1}{|c|}{\multirow{2}{*}{$81.30$}} \\ & & & $92.9$ & $90.0$ & $91.3$ & $96.0$ & $\blue{89.9}$ & $74.5$ & $62.0$ & $92.0$ & $\blue{90.0}$ & & 
    \\ \hdashline
    \multicolumn{1}{|l|}{\multirow{2}{*}{\textbf{\textit{SegFace}}}} & \multicolumn{1}{|c|}{\multirow{2}{*}{AAAI 25}} & \multicolumn{1}{|c|}{\multirow{2}{*}{$512$}} & $\blue{96.7}$ & $\blue{94.2}$ & $\blue{95.4}$ & $\blue{90.9}$ & $\blue{91.1}$ & $\blue{87.2}$ & $\blue{87.1}$ & $\blue{89.3}$ & $\blue{88.9}$ & \multicolumn{1}{|c|}{\multirow{2}{*}{$\blue{88.96}$}} & \multicolumn{1}{|c|}{\multirow{2}{*}{$\blue{81.55}$}} \\ & & & $\blue{93.1}$ & $\blue{90.3}$ & $\blue{91.6}$ & $\blue{96.0}$ & $89.3$ & $\blue{75.1}$ & $\blue{63.4}$ & $\blue{92.1}$ & $89.8$ & & 
    \\ \hline
\end{tabular}
}
\\(b) CelebAMask-HQ dataset 
\caption{Quantitative results on (a) LaPa dataset and (b) CelebAMask-HQ dataset}
\label{table:results}
\vspace{-1em}
\end{table*}

\subsection{Output Head} The output head's role is to generate the final segmentation maps from the learned class-specific tokens and the updated face tokens. The face tokens \( F' \) are upscaled using a small network \( h_{\alpha} \), which comprises transpose convolution operations. The upscaling increases the resolution of the face tokens to match the original image size. Formally, this can be defined as \( U = h_{\alpha}(F') \), where \( U \in \mathbb{R}^{B \times C' \times H \times W} \) is the upscaled face token embedding, and \( C' \) is the reduced embedding dimension after upscaling. Finally, the learned class-specific tokens \( \mathbb{\tilde{T}} = \{\tilde{T}_i \mid 1 \leq i \leq N\} \) are passed through an MLP and then multiplied element-wise with the upscaled face tokens to produce the final segmentation maps:
\[
S_i = U \odot \text{MLP}(\tilde{T}_i),
\]
where \( \odot \) denotes element-wise multiplication, and \( S_i \in \mathbb{R}^{B \times 1 \times H \times W} \) represents the segmentation map for the \( i \)-th class. The final output is a set of segmentation maps \( \mathbb{S} = \{S_i \mid 1 \leq i \leq N\} \) for all classes, where each \( S_i \) corresponds to a specific face component, effectively segmenting the input face image into its respective regions.

\section{Experiments}
\begin{figure*}[t]
    \centering
    \includegraphics[width=\textwidth]{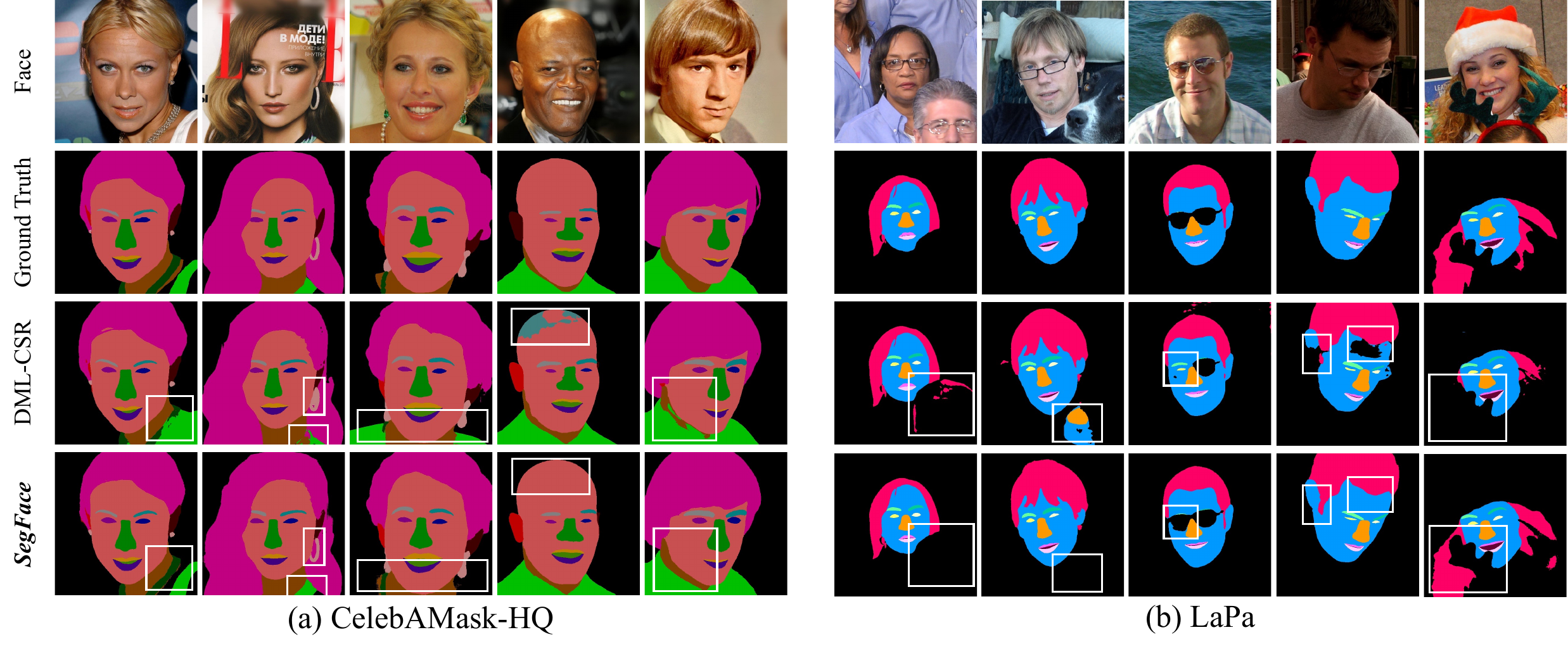}
    \caption{The qualitative comparison highlights the superior performance of our method, \textit{SegFace}, compared to DML-CSR. In (a), \textit{SegFace} effectively segments both long-tail classes like earrings and necklaces as well as head classes such as hair and neck. In (b), it also excels in challenging scenarios involving multiple faces, human-resembling features, poor lighting, and occlusion, where DML-CSR struggles.}
    \label{fig:qualitative_results}
    \vspace{-1em}
\end{figure*}
\subsection{Datasets}
We conduct our experiments on three standard face segmentation datasets: LaPa~\cite{liu2020new}, CelebAMask-HQ~\cite{Lee_2020_CVPR}, and Helen~\cite{le2012interactive}. The LaPa dataset contains a total of 22,168 images, with 18,176 used for training, 2,000 for validation, and 2,000 for testing. This dataset is annotated for 11 classes, including skin, hair, nose, left eye, right eye, left brow, right brow, upper lip, and lower lip. The CelebAMask-HQ dataset comprises 30,000 face images, split into 24,183 for training, 2,993 for validation, and 2,824 for testing. It features 19 semantic classes, including accessories such as earring, necklace, eyeglass, and hat, which are considered long-tail classes due to their infrequent occurrence in the dataset. The other classes are the same as those in the LaPa dataset, with the addition of left/right ear, cloth and neck. The Helen dataset, being the smallest, consists of 2,000 training samples, 230 validation samples, and 100 test samples, annotated for 11 classes.

\subsection{Implementation Details}
We trained \textit{SegFace} in various configurations by changing the backbones (Swin, Swin V2, ResNet101, MobileNetV3, EfficientNet) and input resolutions ($64$, $96$, $128$, $192$, $224$, $256$, $448$, $512$). The models were optimized for 300 epochs using the AdamW optimizer, with an initial learning rate of $1e^{-4}$ and a weight decay of $1e^{-5}$. We employed a step LR scheduler with a gamma value of $0.1$, which reduces the learning rate by a factor of $0.1$ at epochs $80$ and $200$. A batch size of $32$ was used for training on the LaPa and CelebAMask-HQ datasets, and $16$ for the Helen dataset. We did not perform any augmentations on the CelebAMask-HQ and Helen datasets. For the LaPa dataset, we applied random rotation $[-30^\circ, 30^\circ]$, random scaling $[0.5, 3]$, and random translation $[-20\text{px}, 20\text{px}]$, along with RoI tanh warping~\cite{lin2019face} to ensure that the network focused on the face region. The $\lambda_1$ and $\lambda_2$ values were set at $0.5$ for dice loss and cross entropy loss, respectively. Our method was evaluated against other baselines using class-wise F1 score, mean F1 score, and mean IoU, with the background class excluded in all metrics. All code was implemented in PyTorch, and the models were trained on eight A6000 GPUs, each equipped with $48$ GB of memory.

\section{Results and Analysis}
In this section, we detail the quantitative and qualitative results of \textit{SegFace} and demonstrate its superiority in handling the segmentation of long-tail classes. Further, we analyze the benefits of the proposed method. 

\textbf{Quantitative Results:} The class-wise F1-score, mean F1-score, and mean IoU on the LaPa and CelebAMask-HQ datasets are shown in Table~\ref{table:results}(a) and Table~\ref{table:results}(b), respectively. We observe that \textit{SegFace} outperforms other existing methods, achieving a mean F1-score of $93.03$ and a mean IoU of $88.14$ on the LaPa dataset. We see improvements in majority of the classes, with the largest gains in the lower-lip, inner-mouth, and upper-lip classes, with increments of $0.6$, $0.7$, and $0.7$, respectively. The performance improvement in these classes validates our claim that multi-scale feature extraction and fusion help mitigate the scale-discrepancy problem between different facial regions, thereby boosting overall segmentation performance. \textit{SegFace} also significantly outperforms other baselines on the CelebAMask-HQ dataset, achieving a mean F1-score of $88.96$ ($+2.89$) and a mean IoU of $81.55$ ($+3.74$). Specifically, we observe significant improvements in the long-tail classes such as eyeglasses, earrings, and necklaces, with increments of $2.8$, $5.5$, and $22.8$, respectively. In addition to these improvements in long-tail classes, \textit{SegFace} also shows enhanced performance across other classes in the CelebAMask-HQ dataset, outperforming other methods when comparing the class-wise F1 score. This significant performance improvement can be attributed to the transformer decoder with learnable class-specific tokens. It associates each class with a specific token and prevents the dominance of head classes during training, ensuring effective feature representation for the long-tail classes. Additionally, the cross-attention between fused features and tokens helps the tokens extract class-specific information and enables independent modeling of classes.

\textbf{Qualitative Results:} We illustrate the qualitative comparison of our proposed method against other baselines in Figure~\ref{fig:qualitative_results}. From Figure~\ref{fig:qualitative_results}(a) [columns 1,2,3], we validate that \textit{SegFace} is capable of segmenting long-tail classes such as earring and necklace much better compared to the existing state-of-the-art method, DML-CSR. This demonstrates the effectiveness of the proposed transformer decoder with learnable task-specific queries. It enables independent modeling of all classes by associating each token with a particular class. In this design, the token can focus specifically on that class and learn to leverage the fused features for segmentation. Furthermore, from Figure~\ref{fig:qualitative_results}(a) [columns 4,5], we observe that the proposed method also performs better on head classes such as hair and neck.  The results on the LaPa dataset, as shown in Figure~\ref{fig:qualitative_results}(b) [columns 1, 2], indicate that DML-CSR struggles with face segmentation in the presence of multiple faces or human-resembling features in the vicinity. We mitigate this issue by incorporating RoI Tanh warping~\cite{lin2019face} to ensure that the model focuses on the face region while performing segmentation. From Figure~\ref{fig:qualitative_results}(b) [columns 3,4], we can see that DML-CSR performs poorly in challenging lighting conditions and in Figure~\ref{fig:qualitative_results}(b) [column 5], it struggles with occlusion. \textit{SegFace} outperforms DML-CSR and is able to accurately segment facial regions even in these complex scenarios.  

\begin{figure}[ht]
    \centering
    \includegraphics[width=\linewidth]{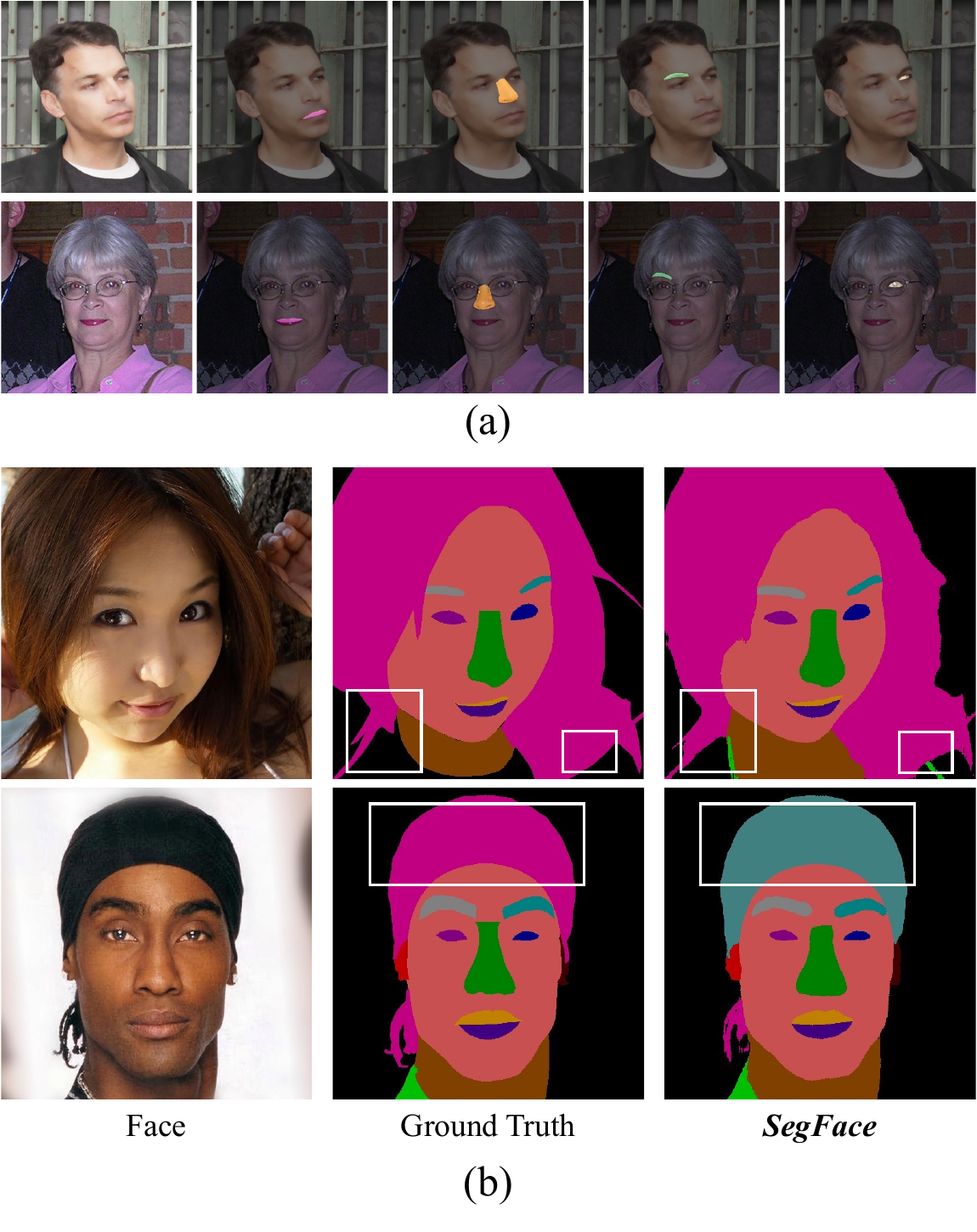}
    \caption{(a) Class-specific tokens segment their corresponding classes, showcasing the independent modeling of each class. (b) Comparison of noisy ground truth with prediction from \textit{SegFace}}
    \label{fig:viz}
    \vspace{-1em}
\end{figure}

\textbf{Analysis:} We make the following claims: ``The transformer decoder with learnable class-specific queries enables independent modeling of classes" and ``In our proposed approach, each token is associated with one class, allowing it to focus specifically on that particular class." To validate these claims, we analyze what each token is learning. We visualize the segmentation outputs of some tokens such as upper-lip, nose, left-brow and right-eye in Figure~\ref{fig:viz}(a). We observe that each token effectively learns the class it has been associated with, demonstrating independent modeling of classes. The learnable tokens leverage the shared fused features via cross-attention to learn the class-specific information. Furthermore, we manually analyzed the segmentation outputs and compared them with the ground truth. We found that the proposed approach provides accurate segmentation output even in the presence of samples with noisy ground truths, showcasing its robustness. The noisy ground truths and our predictions for the same are illustrated in Figure~\ref{fig:viz}(b).

\vspace{-0.3em}
\section{Ablation Studies}
We conduct an ablation analysis to study different components in our proposed approach and provide helpful insights.

\textbf{Varying the backbone of \textit{SegFace}:} We trained \textit{SegFace} with various backbones to demonstrate the strength of the proposed lightweight transformer decoder with learnable task-specific tokens. As shown in Table~\ref{table:ablation}(a), we conducted experiments using backbones with parameter sizes ranging from $7$M to $91$M and observed that the segmentation performance remained consistent with minimal variation. This consistency indicates that the transformer decoder is responsible for majority of the heavy lifting, making it the core component of our proposed approach. Furthermore, we want to emphasize that the proposed method can be adapted for low-compute edge devices by simply swapping the backbone to MobileNetV3~\cite{DBLP:journals/corr/abs-1905-02244}. The mobile version achieves $95.96$ FPS with a mean F1 score of $87.91$ ($+1.77$) on the CelebAMask-HQ dataset, surpassing the current state-of-the-art. 

\textbf{\textit{SegFace} w/o multi-scale feature extraction:} We trained \textit{SegFace} using the single-scale final feature obtained from the backbone without any feature fusion, as shown in Table~\ref{table:ablation}(a) [Row 5]. As expected, we observed a drop in performance when the model was trained without multi-scale feature extraction. This showcases the importance of multi-scale feature extraction and feature fusion in effectively handling different face regions that appear at varying scales. 

\textbf{Performance at different input resolutions:}
We analyzed the performance and FPS of \textit{SegFace} at different input resolutions to showcase the trade-off between FPS and performance, which can be valuable for applications requiring lower memory usage and inference costs. Notably, \textit{SegFace}, even when trained at a low-resolution of $192 \times 192$, outperforms the best version of current state-of-the-art DML-CSR, which is trained at $512 \times 512$ resolution.
\setlength\dashlinedash{2pt}
\setlength\dashlinegap{3pt}
\setlength\arrayrulewidth{0.4pt}

\begin{table}[t]
\centering
\resizebox{.47\textwidth}{!}{
\renewcommand{\arraystretch}{1.15}
\begin{tabular}{|l|c|c|c|c|}
    \hline
    \textbf{Backbone} & \textbf{Mean F1} & \textbf{Mean IoU} & \textbf{FPS} & \textbf{Params} \\
    \hline
    ResNet 100 & 87.50 & 79.65 & 67.81 & 47.254 \\
    EfficientNet & 88.94 & 81.49 & 40.14 & 57.035 \\
    MobileNet V3 & 87.91 & 79.98 & 95.96 & 7.034 \\
    Swin V2 & 88.73 & 81.30 & 34.13 & 91.168 \\
    Swin (w/o fusion) & 87.83 & 80.07 & 40.00 & 90.513 \\
    Swin & 88.96 & 81.55 & 38.95 & 91.006 \\
    \hline
\end{tabular}
}
\\ (a) \textit{SegFace} performance with different backbones
\\\textcolor{white}{a}\\
\resizebox{.47\textwidth}{!}{
\renewcommand{\arraystretch}{1.15}
\begin{tabular}{|l|c|c|c|c|c|c|c|c|}
    \hline
    \textbf{Res} & \textbf{64} & \textbf{96} & \textbf{128} & \textbf{192} & \textbf{224} & \textbf{256} & \textbf{448} & \textbf{512} \\
    \hline
    FPS & 54.56 & 54.11 & 45.77 & 47.39 & 47.72 & 42.78 & 44.53 & 38.95 \\
    Mean F1 & 80.92 & 83.75 & 85.62 & 87.11 & 87.47 & 87.66 & 88.77 & 88.96 \\
    Mean IoU & 71.72 & 75.20 & 77.24 & 79.18 & 79.65 & 79.91 & 81.30 & 81.55 \\
    \hline
\end{tabular}
}
\\(b) \textit{SegFace} performance for varying image resolutions 
\caption{Ablation study for different backbones and varying image resolution.}
\label{table:ablation}
\end{table}

\setlength\dashlinedash{2pt}
\setlength\dashlinegap{3pt}
\setlength\arrayrulewidth{0.4pt}

\begin{table}[!t]
\centering
\resizebox{0.47\textwidth}{!}{
\renewcommand{\arraystretch}{1.15}
\begin{tabular}{|l|cccccccc|c|}
    \hline
    \multicolumn{1}{|l|}{Method} & \multicolumn{1}{c}{Skin} & \multicolumn{1}{c}{Nose} & \multicolumn{1}{c}{U-Lip} & \multicolumn{1}{c}{I-Mouth} & \multicolumn{1}{c}{L-Lip} & \multicolumn{1}{c}{Eyes} & \multicolumn{1}{c}{Brows} & \multicolumn{1}{c|}{Mouth} & \multicolumn{1}{c|}{Mean F1 $\uparrow$} \\ \hline
    DML-CSR & $96.6$ & $95.5$ & $87.6$ & $91.2$ & $91.2$ & $90.9$ & $88.5$ & $95.9$ & $93.8$ \\
    FP-LIIF & $95.1$ & $94.0$ & $79.7$ & $86.3$ & $87.6$ & $89.1$ & $81.0$ & $93.6$ & $91.2$ \\
    \textit{\textbf{SegFace}} & $95.6$ & $94.5$ & $81.8$ & $87.5$ & $88.7$ & $89.2$ & $83.1$ & $94.4$ & $91.8$ \\
    \hline
\end{tabular}
}
\caption{Results on Helen Dataset}
\label{table:helen}
\vspace{-0.8em}
\end{table}

\section{Conclusion}
In this work, we present \textit{SegFace}, a systematic approach that leverages a lightweight transformer decoder with learnable task-specific tokens to address the challenge of poor segmentation performance on long-tail classes. We also incorporate multi-scale feature extraction and MLP fusion in our pipeline to resolve the scale discrepancy problem between different face regions. Through extensive experiments, we validate the effectiveness of our approach and provide insightful comments to highlight its superiority. The results demonstrate that we significantly outperform other methods, achieving state-of-the-art segmentation performance on the LaPa and CelebAMask-HQ datasets.

\section{Limitation} 
Transformers typically require large amounts of data for optimal training and demonstrate improved performance as the data scales~\cite{DBLP:journals/corr/abs-2005-14165}. \textit{SegFace} leverages a transformer-based decoder and, therefore, exhibits below-SOTA performance with scarce training data, which is its primary limitation. We trained \textit{SegFace} on the Helen dataset, which comprises of  only $2000$ training samples, and summarized the results in Table~\ref{table:helen}.

\section{Acknowledgement}
This research is based upon work supported in part by the Office of the Director
of National Intelligence (ODNI), Intelligence Advanced Research Projects Ac-
tivity (IARPA), via [2022-21102100005]. The views and conclusions contained
herein are those of the authors and should not be interpreted as necessarily
representing the official policies, either expressed or implied, of ODNI, IARPA,
or the U.S. Government. The US Government is authorized to reproduce and
distribute reprints for governmental purposes notwithstanding any copyright an-
notation therein.

\bibliography{aaai25}

\newpage
\section*{Appendix}
\begin{figure*}[t]
    \centering
    \includegraphics[width=\textwidth]{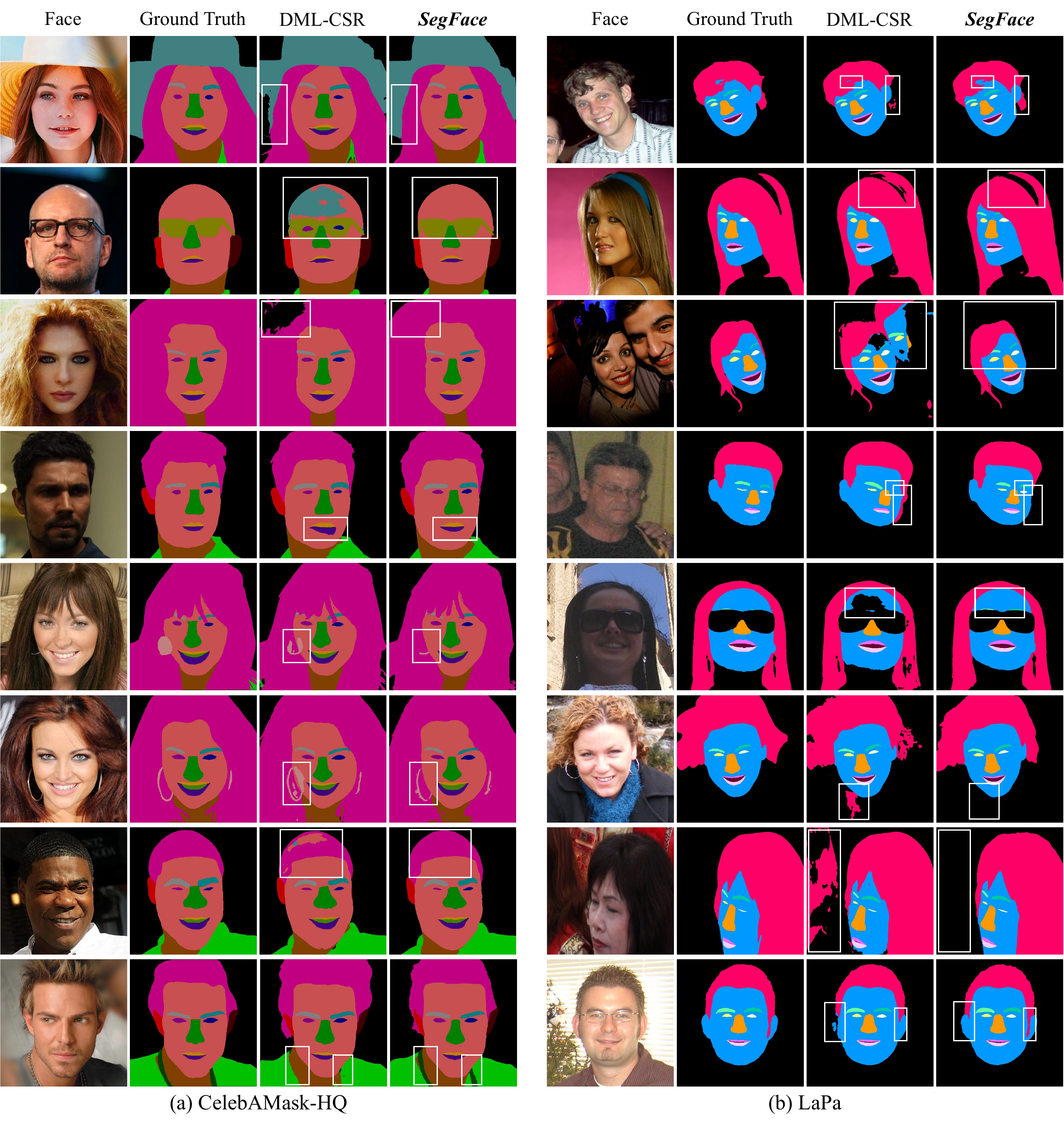}
    \caption{Additional qualitative comparison of our proposed method, \textit{\textbf{SegFace}}, compared to DML-CSR on the (a) CelebAMask-HQ and (b) LaPa dataset.}
    \label{fig:supp_viz}
\end{figure*}

In Appendix, we present an additional qualitative comparison between our proposed method, \textit{\textbf{SegFace}}, and DML-CSR, the current state-of-the-art face parsing model. The results are illustrated in Figure~\ref{fig:supp_viz}.

Focusing first on the visualization of the (a) CelebAMask-HQ dataset, we observe that \textit{SegFace} demonstrates superior performance on long-tail classes such as earring (row 6), necklace (row 8), and hat (row 1). Additionally, it performs better on head classes such as the lower-lip (row 4) and hair (rows 2, 3, 7). \textit{SegFace} also provides accurate segmentation even in the presence of noisy ground truths (row 5).

Shifting our focus to the (b) LaPa dataset, \textit{SegFace} delivers better hair segmentation performance compared to DML-CSR (rows 1, 2, 4, 8). \textit{SegFace} effectively segments similarly textured features like hair and fur, which DML-CSR often confuses (row 6). It also achieves better segmentation performance for classes like skin (row 5), the right brow (row 4), and the right eye (row 4). Moreover, \textit{SegFace} maintains precise segmentation even when people are present in the background or at the edges, where DML-CSR struggles (rows 3, 7).
 
\end{document}